 \documentclass[pmlr,twocolumn]{jmlr} 

\usepackage{booktabs}
\usepackage[load-configurations=version-1]{siunitx} 
 \usepackage{todonotes}
 
 \usepackage{booktabs,arydshln}
 
 \makeatletter
\def\adl@drawiv#1#2#3{%
        \hskip.5\tabcolsep
        \xleaders#3{#2.5\@tempdimb #1{1}#2.5\@tempdimb}%
                #2\z@ plus1fil minus1fil\relax
        \hskip.5\tabcolsep}
\newcommand{\cdashlinelr}[1]{%
  \noalign{\vskip\aboverulesep
           \global\let\@dashdrawstore\adl@draw
           \global\let\adl@draw\adl@drawiv}
  \cdashline{#1}
  \noalign{\global\let\adl@draw\@dashdrawstore
           \vskip\belowrulesep}}
\makeatother

\theorembodyfont{\upshape}
\theoremheaderfont{\scshape}
\theorempostheader{:}
\theoremsep{\newline}

\title[Oral Cancer Detection]{Transfer Learning for Oral Cancer  \titlebreak Detection  using Microscopic Images}

\author{%
\Name{Rutwik Palaskar} \Email{rutwikpalaskar@gmail.com} \\
\addr School of Bioengineering Sciences and Research, MIT ADT University
\AND
\Name{Renu Vyas} \Email{renu.vyas@mituniversity.edu.in} \\
\addr School of Bioengineering Sciences and Research, MIT ADT University
\AND
\Name{Vilas Khedekar} \Email{vilaskhedekar2010@gmail.com}\\
\addr School of Bioengineering Sciences and Research, MIT ADT University
\AND
\Name{Sangeeta Palaskar} \Email{palaskarsangeeta@gmail.com}\\
\addr Department of Oral and Maxillofacial Pathology, Sinhgad Dental College and Hospital
\AND
\Name{Pranjal Sahu} \Email{psahu@cs.stonybrook.edu}\\
\addr Department of Computer Science, Stony Brook University
}

\begin{document}

\maketitle

\begin{abstract}







Oral cancer has more than 83\% survival rate if detected in its early stages, however, only 29\% cases are currently  detected early. Deep learning techniques can detect patterns of oral cancer cells and can aid in its early detection. In this work, we present first results of neural networks for oral cancer detection using microscopic images. We compare numerous state-of-the-art models via transfer learning approach, and collect and release an augmented dataset of high-quality microscopic images of oral cancer. We present a comprehensive study of different models and report their performance on this type of data. Overall, we obtain a 10-15\% absolute improvement with transfer learning methods compared to a simple Convolutional Neural Network baseline. Ablation studies show the added benefit of data augmentation techniques with finetuning for this task. 



\end{abstract}

\section{Introduction}
\label{sec:intro}
Oral cancer is the 6$^{th}$ most common cancer in the world \citep{warnakulasuriya2009global}. According to the World Health Organisation, there are an estimated 657,000 new cases of cancers of the oral cavity and pharynx each year and more than 330,000 deaths worldwide \citep{who_study}. Mortality rate is very high for oral cancer due to lack of early diagnosis. 

\begin{figure}[t]
\centering
\includegraphics[width=\linewidth]{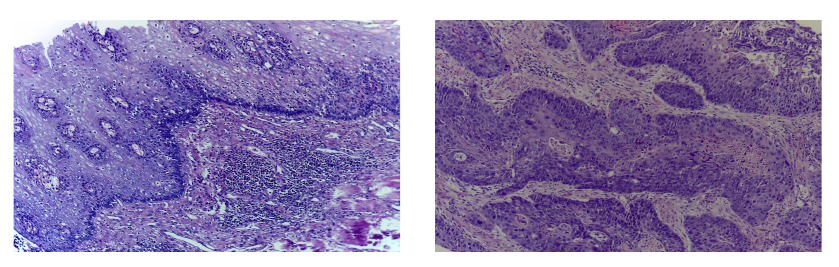}
\caption{Example of benign (left) and malignant (right) oral squamous cells.}
\end{figure}

Oral cancer develops in the tissues of the mouth or throat, and most develop in the squamous cells in the cheeks, tongue and lips \citep{thompson2006world,epstein2010oral}. According to the National Cancer Research Institute, the five year survival rates for oral cavity and pharynx cancer are: 83\% for localized cancer (that has not spread), 64\% for cancer that has spread to nearby lymph nodes, and 38\% for cancer that has spread to other parts of the body \citep{nih_study}.

Based on the above facts, early diagnosis is very crucial for oral cancer. Early cancer detection and prognosis is one of the several important healthcare areas where deep learning technologies have been applied \citep{kourou2015machine}. It has shown huge success in diagnosing various medical conditions with high accuracy, specificity and sensitivity \citep{rajpurkar2017chexnet,abramoff2016improved}.

At the microscopic levels, malignant squamous cells are larger in size than that of the normal cells and are distinctly different from each other in their shapes \citep{ahmed2019oral}. Confirmatory diagnosis of oral cancer is done by its microscopic examination which needs to be done by a highly qualified and experienced specialist. Automating this examination process can significantly reduce the burden of specialists, however, very few studies have been done for oral cancer diagnosis by using algorithm guided detection at the microscopic levels. Towards this goal, we propose to use transfer learning techniques from large-scale image classification applied to medical image classification, for classifying oral squamous cells as cancerous or not at the microscopic levels.


In this paper, for the first time, we share the performance of deep learning and transfer learning based models for classifying oral squamous cells as cancerous or non-cancerous. Along with that we perform detailed analysis of model's performance using various metrics with ablation studies for transfer learning and data augmentation to report the best performing model for this task. 

Additionally, we also collect and release original data of normal oral squamous cells and cancerous oral squamous cells in collaboration with a Dental hospital to augment the currently available dataset (there is only one such dataset available currently), and also to report multi-site testing results on this type of data. Further, class-imbalance being a well known problem for medical image classification, we evaluate three commonly used techniques to address this.

\begin{table*}[t]
  \centering
  \begin{tabular}{llcccccc}
    \toprule
    \textbf{No.} & \textbf{Model} & \textbf{\# Params}    & \textbf{Acc}    & \textbf{P}    & \textbf{R}    & \textbf{F1}    & \textbf{AUC} \\
    \midrule
    1     	& Small CNN  & 16M & 75.81 & 0.38 & 0.50 & 0.43 & 0.5 \\
    2       & Large CNN & 67M & 76.61 & 0.58 & 0.52 & 0.48 & 0.52\\
    \midrule
    3       & MobileNet        & 2K & 77.42 & 0.69 & 0.69 & 0.69 & 0.69 \\
    3.1        & \quad \& Data Aug & 2K & 79.84 & 0.72 & 0.71 & 0.72 & 0.71\\
    3.2        & \quad \& FT & 2M & \textbf{85.48} & \textbf{0.81} & \textbf{0.79} & \textbf{0.80} & \textbf{0.79}  \\
    3.3        & \quad \& Data Aug + FT     & 2M & 83.87 & 0.79 & 0.75 & 0.76 & 0.75 \\
    \midrule
    4       & ResNet        & 4K & 83.06 & 0.91 & 0.65 & 0.68 & 0.65 \\
    4.1        & \quad \& Data Aug & 4K & 79.03 & 0.74 & 0.61 & 0.63 & 0.61 \\
    4.2        & \quad \& FT & 23M & 87.9 & 0.86 & 0.8 & 0.82 & 0.8\\
    4.3       & \quad \& Data Aug + FT     & 23M & \textbf{91.13} & \textbf{0.88} & \textbf{0.87} & \textbf{0.88} & \textbf{0.87} \\
    \midrule
    5       & Inception V3        & 4K & 83.87 & \textbf{0.91} & 0.67 & 0.70 & 0.67 \\
    5.1        & \quad \& Data Aug &  4K & 84.68 & 0.81 & 0.74 & 0.77 & 0.74 \\
    5.2        & \quad \& FT & 21M & 87.1 & 0.82 & \textbf{0.84} & 0.83 & \textbf{0.84} \\
    5.3        & \quad \& Data Aug + FT     & 21M & \textbf{89.52} & 0.89 & 0.82 & \textbf{0.84} & 0.82 \\
    \bottomrule
  \end{tabular}
  \caption{Simple CNNs and transfer learning results using MobileNet, ResNet and Inception V3 with an ablation study on finetuning (FT), data augmentation, and both. We report accuracy (Acc), Precision (P), Recall (R), F1, and AUC-ROC values.}
  \label{tab:main_results}
\end{table*}

\section{Related Work}
\label{sec:related_work}
Transfer learning of large-scale pre-trained image classification models for medical image classification  has been widely adopted in the fields of radiology \citep{rajpurkar2017chexnet,wang2017chestx}, opthalmology \citep{abramoff2016improved,de2018clinically,gulshan2016development,raghu2019direct}, early detection of Alzheimer's \citep{ding2019deep}, identifying skin cancer \citep{esteva2017dermatologist}, and determining embryo quality of IVF procedures \citep{khosravi2018robust}.

Within the head and neck and oral cavity imaging, some research has been on classifying mouth and lip region using photographic images \citep{fu2020deep}, classification of cell types within cancerous cells \citep{das2020automated}, and on stages of oral cancer using Gaussian modeling \citep{rajaguru2017performance}.

\section{Models}
\label{sec:models}

To lay the groundwork of our study, we evaluated two simple Convolutional Neural Network (CNN) architectures. Since transfer learning is proven to be the go-to method for medical imaging, we used MobileNet \citep{howard2017mobilenets}, ResNet \citep{he2016deep,he2019rethinking} and Inception V3 \citep{szegedy2015going} pre-trained models with and without finetuning. Additionally, we also perform data augmentation by rotation, flipping, and shift ranges to synthesize more data.

\paragraph{Simple CNNs} We use two simple CNNs models as baselines to compare with transfer learning models. The Small CNN model is constructed to have only 16M parameters and the Large CNN has 67M parameters. The architecture is shown in Figure \ref{fig:model_architecture} in Appendix \ref{apd:second} with other model details.

\paragraph{Transfer Learning} We use two standard models trained on the common \texttt{ImageNet} dataset \citep{krizhevsky2012imagenet}, ResNet-50 \citep{he2016deep}, and Inception V3 \citep{szegedy2015going}. We also compare with MobileNet \citep{howard2017mobilenets} which is a much lighter model than others.

\section{Data and Experimental Setup}
\label{sec:experimental_setup}
\paragraph{Dataset}
Our dataset was collected from a histopathological image repository of the normal epithelium of the oral cavity and the Oral Squamous Cell Carcinoma (OSCC) images \citep{rahman2020histopathological}. The repository consists of 1224 total images. They are divided into two sets in two different resolutions, 100x magnification and 400x magnification. In total, there are 290 normal epithelium images and 934 OSCC images. We randomly split this data into 10\% validation and testing set each containing same class distribution of normal and cancerous images.

\paragraph{Multi-site Data Collection} Separate from the above repository, we collaborate with a hospital to collect additional images of OSCC to perform multi-site data collection and model testing. Large scale data collection is still is process here. In this work, we used 123 OSCC images collected at this site to test our machine learning models. As this set only contains cancerous images, we augment it with non-cancerous images from the existing test set described above.

\paragraph{Ablation Studies} Within transfer learning, we perform ablation studies to measure impact of data augmentation and fine-tuning by themselves and together. 

\paragraph{Handling Class Imbalance} We follow three common techniques of handling class imbalance \citep{buda2018systematic}: (1) \textit{Oversampling} where images from the minority class are randomly sampled until the cardinality of both classes is same, and equal to the majority class. (2) \textit{Undersampling} randomly samples the majority class to choose fewer images such that the cardinality of both classes is same, and equal to the minority class. (3) \textit{Loss Weighting} does not modify the data but scales the weights for each class during training, encouraging the model to pay more ``attention'' to the minority class.

\paragraph{Evaluation} We perform evaluations across numerous standard classification metrics like accuracy, Precision, Recall, F1, AUC ROC, and analysis via AUC ROC and Confusion Matrix. 


\section{Results and Discussion}
\label{sec:results}

\paragraph{Transfer Learning.} Table \ref{tab:main_results} reports Small and Large CNNs and MobileNet, ResNet and Inception V3 transfer learning results with ablation studies on transfer learning. Overall, we observe best performance on test accuracy and AUC-ROC with Inception V3 finetuned on our dataset. Data augmentation gives added performance in larger models like ResNet and Inception V3. Small and Large CNNs overfit on the majority class leading to AUC-ROC scores around 0.5.


\begin{table}[t]
  \centering
  \begin{tabular}{lccc}
    \toprule
    \textbf{Technique} & \textbf{Acc}    & \textbf{P}    & \textbf{AUC}    \\
    \midrule
    \multicolumn{4}{c}{\texttt{Large CNN}}      \\
    \cdashlinelr{1-4}
    Oversampling   & \textbf{66.13}  & 0.32  & 0.56  \\
    Undersampling  & 28.23 & 0.25  & 0.52 \\
    Loss Weighting  & 51.6 & \textbf{0.33}  & \textbf{0.66} \\
    \midrule
    \multicolumn{4}{c}{\texttt{Inception V3}}      \\
     \cdashlinelr{1-4}
    Oversampling   & \textbf{94.35} & \textbf{0.93} & \textbf{0.91}  \\
    Undersampling  & 75.0 & 0.71 & 0.78  \\
    Loss Weighting  & 83.06 & 0.78  & 0.85 \\
    \bottomrule
  \end{tabular}
  \caption{Comparing class imbalance handling techniques.}
  \label{tab:class_imbalance_results}
\end{table}

\paragraph{Handling Class Imbalance.} Table \ref{tab:class_imbalance_results} reports best results with loss weighting for Large CNN whereas oversampling works best for Inception V3 with finetuning and data augmentation. Following from Table \ref{tab:main_results}, Large CNN overall under-performs compared to Inception V3, hence while loss weighting has a lower accuracy, it has better AUC-ROC value showing better classifier balance across classes.

%



\begin{table}
  \centering
  \begin{tabular}{lccc}
    \toprule
    \textbf{Model} & \textbf{Acc}    & \textbf{P}     & \textbf{AUC}    \\
    \midrule
    Large CNN   & 79.74  & 0.4  &  0.5 \\
    MobileNet & 74.51 & 0.64  & 0.69 \\
    ResNet  & 81.70 & 0.72  & \textbf{0.75} \\
    Inception V3  & \textbf{83.66} & \textbf{0.74} & 0.71 \\
    \bottomrule
  \end{tabular}
  \caption{Multi-site testing results.}
  \label{tab:multi_site_results}
\end{table}

\paragraph{Multi-site Results.} Multi-site testing results follow similar model trends as the training and test data distribution. Table \ref{tab:multi_site_results} shows competitive results with all three transfer learning with finetuning models.

\paragraph{Confusion Matrix Analysis.} Confusion Matrix comparison of True Positive, False Positive, True Negative, False Negatives for Large CNN and Inception V3. Both models recognize the True Positive cases with equal performance but Large CNN performs poorly on True Negatives.

\begin{figure}[t]
\centering
\includegraphics[width=\linewidth]{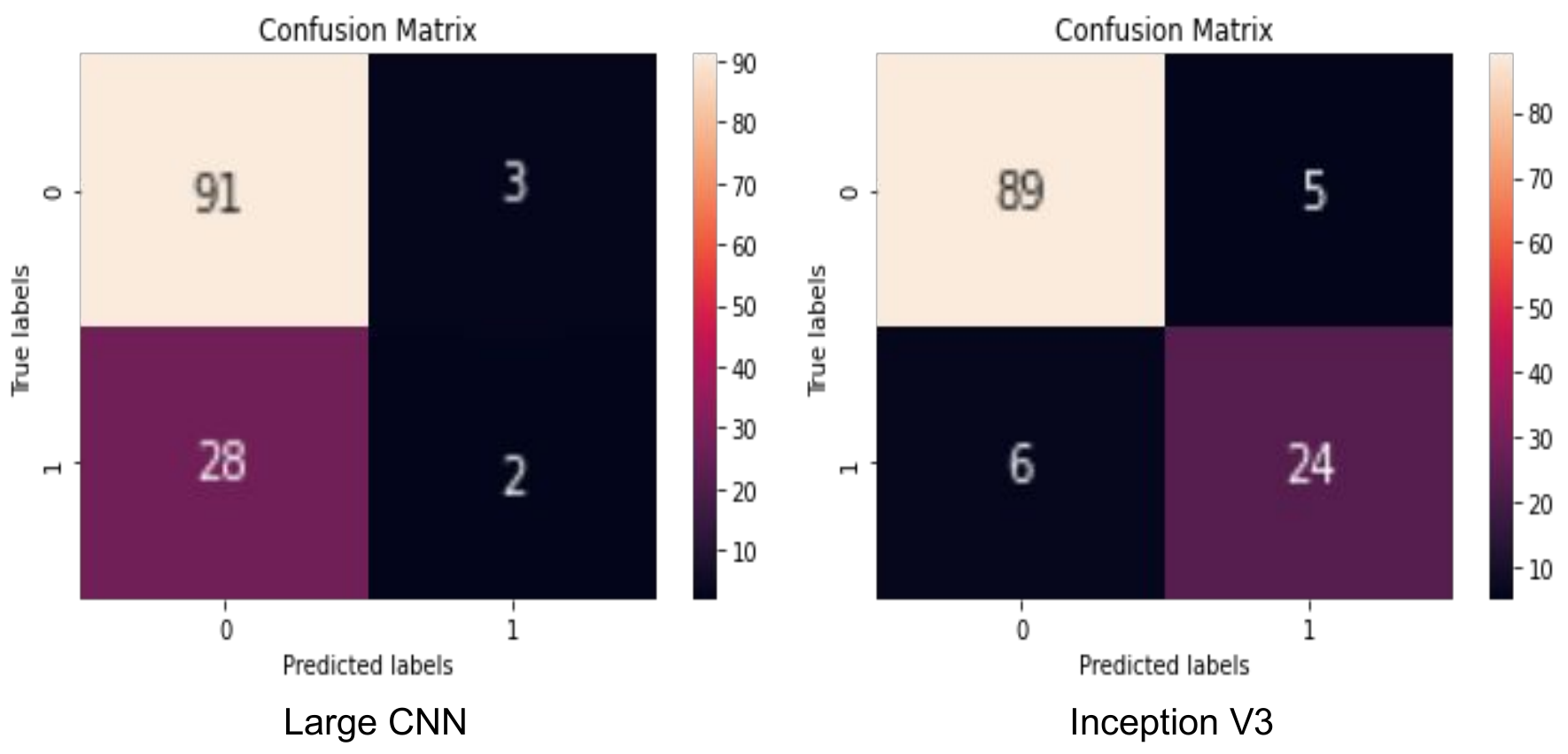}
\caption{Confusion Matrix analysis for Large CNN and Inception V3.}
\end{figure}

\section{Conclusion}
\label{sec:conclusion}
We present first results on classification of the histopathological images of the oral cancer. Using a newly collected repository of such data, we run various state-of-the-art image classification models with transfer learning, finetuning, and data augmentation. We additionally collect more data from a different site to demonstrate multi-site model testing results. We observe strong performance with transfer learning models that lay the groundwork for further research in this type of medical image classification.


\bibliography{jmlr-sample}

\appendix

\section{Dataset Details}\label{apd:data}
\paragraph{Histopathological Image Repo} The repository \citep{rahman2020histopathological} consists of 1224 total images. They are divided into two sets in two different resolutions, 100x magnification and 400x magnification. The first set consists of 89 histopathological images with the normal epithelium of the oral cavity and 439 images of OSCC in 100x magnification. The second set consists of 201 images with the normal epithelium of the oral cavity and 495 histopathological images of OSCC in 400x magnifications. The images were captured using a Leica ICC50 HD microscope from H\&E stained tissue slides collected, prepared, and cataloged by medical experts from 230 patients. We combined all the 1$^{st}$ set and 2$^{nd}$ set images of the normal epithelium of the oral cavity and OSCC images respectively.

\paragraph{Multi-site Data Collection} The clinical image data of patients was collected at the \texttt{Anonymous Hospital}. The research microscope used for this task was Leica DM1000, LAS and the images were captured under 100 and 400 magnification. The slides were stained with Hematoxylin Eosin. These images are confirmed cases of Oral Squamous Cell Carcinoma collected from patients. Per patient one sample was taken to confirm oral squamous cell carcinoma. We used a total of 123 microscopic images to test how our machine learning models performed on a completely new test dataset. 

\section{Model Details}\label{apd:second}
\begin{figure}[t]
\centering
\includegraphics[width=0.7\linewidth]{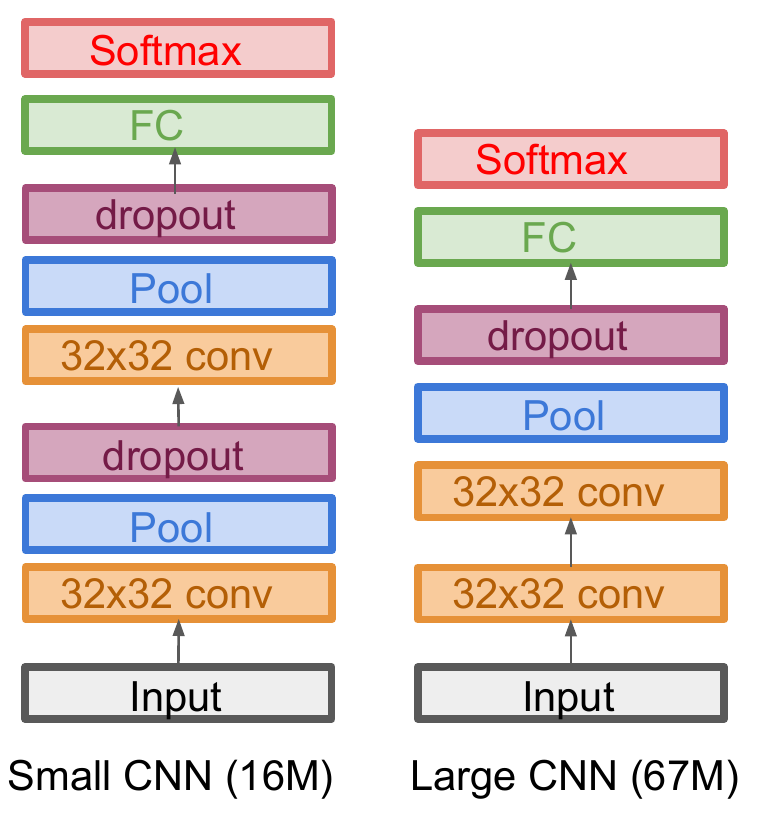}
\label{fig:model_architecture}
\caption{Simple CNN architectures used.}
\end{figure}

\paragraph{Model Details} The first model had the architecture as follows: Two convolution layers, one max pool layer, and one dropout layer. The second model had the architecture as follows: Two convolution, two max pool, and two dropout layers. Due to data imbalance, interesting results were observed with these two models. Figure \ref{fig:model_architecture} shows the model architecture used.

\paragraph{Hyperparameter Settings} We train all models for 10 epochs after observing convergence with loss values on the validation set. Input dimensions for Small and Large CNNs was 256, kernel size was 3, strides 2, and relu activation. We applied a dropout of 0.4. The Fully Connected layer has 128 dimensions.

\end{document}